\pdfoutput=1

\documentclass[11pt]{article}

\usepackage{EMNLP2022}

\usepackage{times}
\usepackage{latexsym}
\usepackage{xspace}
\usepackage{booktabs} 
\usepackage{graphicx}
\usepackage{subcaption}
\usepackage{amstext}
\usepackage{amsmath}
\usepackage{amsfonts}
\usepackage{amsthm}
\usepackage{enumitem}
\usepackage{bigstrut}

\usepackage[T1]{fontenc}

\usepackage[utf8]{inputenc}

\usepackage{microtype}

\usepackage{inconsolata}

\newcommand{\datawebsite}{\textit{ClinicalTrials.gov}\xspace}
\newcommand{\modelname}{\texttt{TrialBERT}\xspace}
\newcommand{\method}{\texttt{Trial2Vec}\xspace}

\newcommand{\bx}{\mathbf{x}}
\newcommand{\bv}{\mathbf{v}}
\newcommand{\bW}{\mathbf{W}}

\title{\method: Zero-Shot Clinical Trial Document Similarity Search using Self-Supervision}

\author{Zifeng Wang and Jimeng Sun \\
 University of Illinois Urbana-Champaign \\
  \texttt{\{zifengw2,jimeng\}@illinois.edu} \\}

\begin{document}
\maketitle
\begin{abstract}
Clinical trials are essential for drug development but are extremely expensive and time-consuming to conduct.  
It is beneficial to study similar historical trials when designing a clinical trial. However, lengthy trial documents and lack of labeled data make trial similarity search difficult. 
We propose a \textit{zero-shot} clinical trial retrieval method, called \method, which learns through self-supervision without the need for annotating similar clinical trials. Specifically, the \textit{meta-structure} of trial documents (e.g., title, eligibility criteria, target disease) along with clinical knowledge (e.g., UMLS knowledge base \footnote{\url{https://www.nlm.nih.gov/research/umls/index.html}}) are leveraged to automatically generate contrastive samples. Besides, \method encodes trial documents considering meta-structure thus producing compact embeddings aggregating multi-aspect information from the whole document. We show that our method yields medically interpretable embeddings by visualization and it gets 15\% average improvement over the best baselines on precision/recall for trial retrieval, which is evaluated on our labeled 1600 trial pairs. In addition, we prove the pretrained embeddings benefit the downstream trial outcome prediction task over 240k trials. \footnote{Software is available at \url{https://github.com/RyanWangZf/Trial2Vec}.}
\end{abstract}

\section{Introduction}
Clinical trials are essential for developing new medical interventions~\cite{fundamentalsofCT}.  Many considerations come into the design of a clinical trial, including study population, target disease, outcome, drug candidates, trial sites, and eligibility criteria, as in Table \ref{tab:mini_example}. It is often beneficial to learn from related clinical trials from the past to design an optimal trial protocol \cite{wang2022artificial}. 
However, accurate similarity search based on the lengthy trial documents is still in dire need.

\begin{table}[t]
  \centering
  \caption{An example of the meta-structure of clinical trial document drawn from \datawebsite. }
  \resizebox{0.5\textwidth}{!}{%
    \begin{tabular}{p{4.96em}|p{19.25em}}
    \toprule
    Title & Effects of Electroacupuncture With Different Frequencies for Major Depressive Disorder \\
    \hline
    Description & Two groups of subjects will be included 55 subjects in electroacupuncture with 2Hz group...\\
    \hline
    Eligibility Criteria & 1. Inclusion Criteria:\newline{}  1.1. Patients suﬀering from MDD in accordance with the diagnostic criteria;\newline{}  1.2. Hamilton Depression Scale score is between 21 and 35 (mild to moderate MDD);...\newline{}2. Exclusion Criteria:\newline{}  2.1 Patients with bipolar disorder;\newline{}  2.2 Patients with schizophrenia or other mental disorders; ...\\
    \hline
    Outcome Measures & 1. Change in anxiety and depression severity measure by Self-rating depression scale\newline{}2. Change in the severity of depression measure by Hamilton depression scale ..\\
    \hline
    Disease & Major Depressive Disorder \\
    \hline
    Intervention & electroacupuncture \\
    \hline
    ...  & ... \\ 
    \bottomrule
    \end{tabular}%
    }
  \label{tab:mini_example}%
\end{table}%

Self-supervision based pretraining has delivered promising performances for many NLP and CV tasks with fine-tuning \cite{devlin2019bert,liu2019roberta,he2021masked,bao2021beit,wang2022medclip}. Nevertheless, we find there was few work on \textit{zero-shot document retrieval} as most address document retrieval in a supervised fashion \cite{humeau2019poly,khattab2020colbert,guu2020realm,karpukhin2020dense,lin2020distilling,luan2021sparse,wang2021online,hofstatter2020local,li2020sentence,zhan2021optimizing,hofstatter2021intra,hofstatter2021efficiently,jiang2022promptbert} or improve document pre-training for further supervision \cite{beltagy2020longformer,zaheer2020big,ainslie2020etc,zhang2021poolingformer}.

Recently, a burgeoning body of research  \cite{gao2021simcse,wu2021esimcse,wang2022sncse} proposes to execute self-supervised learning to train semantic-meaningful \textit{sentence} embeddings free of labels. However, there are still challenges to apply them for document similarity search:
\begin{itemize}[leftmargin=*, itemsep=0pt, labelsep=5pt]
\item \textbf{Lengthy documents.} These zero-shot BERT retrieval methods all work on short sentences (usually below 10 words) similarity search while trial documents are often above 1k words. Simply encoding lengthy trials by truncating and averaging embeddings of all remaining tokens inevitable leads to poor retrieval quality.
\item \textbf{Inefficient contrastive supervision.} These unsupervised methods take simple instance discriminative contrastive learning (CL) within batch, e.g., SimCSE \cite{gao2021simcse} takes one sentence into the encoder twice to get the positive pairs and all other sentences as the negative. This paradigm has low supervision efficiency  to require a large batch size, large data, and long training time, which is infeasible for learning from long trial documents.
\end{itemize}

In this work, we propose Clinical \textbf{Trial} \textbf{TO} \textbf{Vec}tors, \method, a zero-shot trial document similarity search using self-supervision. We design a trial encoding framework considering the meta-structure to rid the risk that semantic meaning vanishes due to the uniform average of token embeddings. Meanwhile, the meta-structure is utilized to generate contrastive samples for efficient supervision. Medical knowledge is introduced to further enhance the negative sampling for CL. Our main contributions are:
\begin{itemize}[leftmargin=*, itemsep=0pt, labelsep=5pt]
\item We are the first to study the trial-to-trial retrieval task by proposing a label-free SSL model which is able to encode long trials into semantic meaningful embeddings without labels.
\item We propose a data-efficient CL method on medical knowledge and trial meta-structure, which is promising to be extended to further zero-shot structured document retrieval.
\item We demonstrate the superiority of \method on a trial relevance dataset of 1600 trials annoated by domain experts. Also, we show \method can assist better downstream trial outcome prediction on a dataset of 240k trials.
\end{itemize}

\section{Related works}
\subsection{Text \& document retrieval}
\textbf{General texts.} Early information retrieval methods depend on manual engineering \cite{robertson2009probabilistic, yang2017anserini}. By contrast, dense retrieval methods based on distributional word representations, e.g., Word2Vec \cite{mikolov2013efficient}, Glove \cite{pennington2014glove}, Doc2Vec \cite{le2014distributed}, etc., become popular crediting to their superior performance. The advent of deep models, especially the contexualized encoders like BERT \cite{devlin2019bert}, encourages an explosion of neural retrieval methods \cite{van2016learning,zamani2018neural,guo2016deep,dehghani2017neural,onal2018neural,reimers2019sentence,chang2019pre,nogueira2019passage,chen2021co,lin2020distilling,xiong2020approximate,karpukhin2020dense,yates2021pretrained}. However, most of them are based on supervised training on sentence pairs from general texts, e.g., SNLI \cite{bowman2015large}. When label is expensive to acquire, as in the clinical trial case, we need zero-shot learning models. Although, there arose some works to perform post-processing on pretrained BERT embeddings to improve their retrieval quality \cite{li2020sentence,su2021whitening}, their performances are far from optimal without specific training.

\noindent \textbf{Clinical trials.}
Traditional clinical trial query search systems \cite{tasneem2012database,tsatsaronis2012ponte,jiang2014cross,park2020interactive} are established on protocol databases. Contrast to dense retrieval, these methods rely on entity matching with rules thus not flexible enough. Recent works \cite{roy2019towards,rybinski2020clinical,rybinski2021science2cure} propose supervised neural ranking for clinical trial query search. However, all of them work on matching trial titles or relevant segments with an input user query. While \method can also assist query search, it is the first to encode complete trial documents for the trial-level similarity search.

\begin{figure*}[t]
    \centering
    \includegraphics[width=0.85\linewidth]{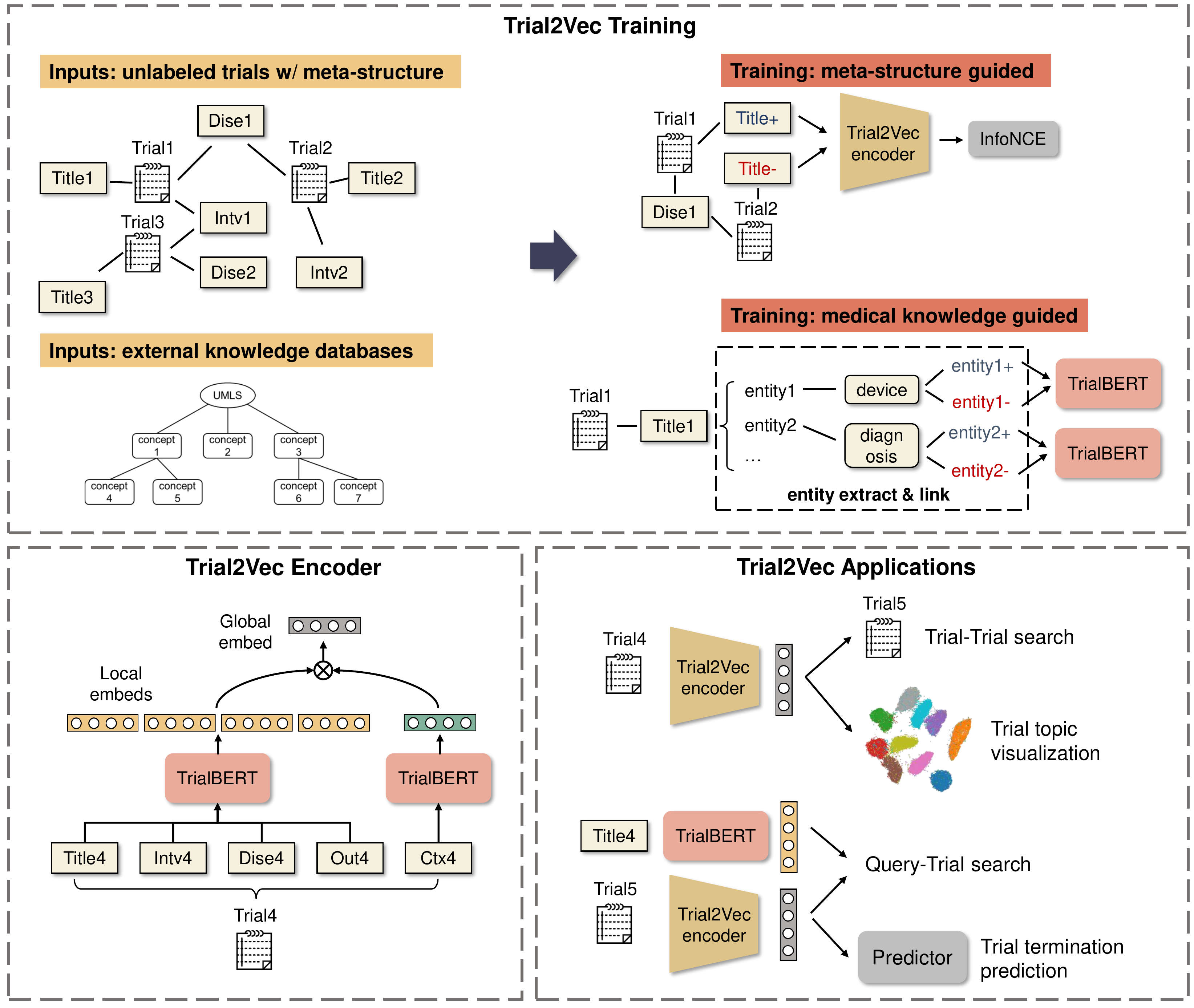}
    \caption{Overview of the proposed \method framework. \textbf{Top left}: the training strategy that accounts for unlabeled input trial documents with meta-structure along with an external medical knowledge database, e.g., UMLS. \textbf{Top right}: The contrastive supervision splits into meta-structure and knowledge guided, respectively. \textbf{Bottom left:} our method hierarchically encodes trials into local and global embeddings on the trial meta-structure. \textbf{Bottom right:} The encoded trial-level embeddings can be used to trial search, query trial search and downstream tasks. \label{fig:overview}}
\end{figure*}

\subsection{Text contrastive learning}
Contrastive learning is a heated discussed topic recently in NLP and CV \cite{chen2020simple,chen2020big,chen2021exploring,carlsson2020semantic,zhang2020unsupervised,wu2020clear,yan2021consert,gao2021simcse,wang2020information,wang2022transtab}. CL is one main topic under the SSL domain. It sheds light on reaching comparable performance as supervised learning free of manual annotations. While CL has been applied to enhance downstream NLP applications like text classification \cite{li2021selfdoc,zhang2022metadata}, a few \cite{wang2020cross,zhang2020unsupervised,yan2021consert,yang2021universal} are able to do zero-shot retrieval. Nonetheless, all focus on enhancing \textit{sentence} embeddings by manipulating text only therefore are suboptimal when facing lengthy documents. By contrast, \method uses the document meta-structure with domain knowledge to obtain and facilitate \textit{document} embeddings.

\section{Method}
In this section, we present the details of \method. The main idea is to jointly learn the global and local representations from trial documents considering their meta-structure. Specifically, observed in Table \ref{tab:mini_example}, trial document consists of multiple sections while the \textit{key attributes} (e.g., title, disease, intervention, etc.) occupy a small portion of the whole document. This motivates us to design a hierarchical encoding and the corresponding contrastive learning framework. The overview is illustrated in Fig. \ref{fig:overview}. Our method generates \textit{local attribute embeddings} using the \modelname backbone separately, then aggregating local embeddings with a learnable attention module to obtain the \textit{global trial embeddings} that emphasize significant attributes. We present the pretraining of backbone encoder in \S \ref{sec:backbone}; then we describe the hierarchical encoding process based on the backbone encoder in \S \ref{sec:trial2vec_encode}; the hierarchical constrastive learning methods considering meta-structure and medical knowledge are elucidated in \S \ref{sec:contrastive_learning}; at last, we elicit the applications of the proposed framework in \S \ref{sec:application}.

\subsection{Backbone encoder: \modelname}\label{sec:backbone}
We leverage the BERT architecture as the backbone encoder in the framework. In detail, we use the WordPiece tokenizer together with the BioBERT \cite{lee2020biobert} pretrained weights as the start point. We continue the pretraining with Masked Language Modeling (MLM) loss on three trial-related data sources: ClinicalTrial.gov \footnote{\url{https://clinicaltrials.gov/}}, Medical Encyclopedia \footnote{\url{https://medlineplus.gov/encyclopedia.html}}, and Wikipedia Articles \footnote{\url{https://www.wikipedia.org/}}, see Table \ref{tab:pretraining_corpus}, to get \modelname. ClinicalTrials.gov is a database that contains around 400k clinical trials conducted in 220 countries. Medical Encyclopedia has 4K high-quality articles introducing terminologies in medicine. We also retrieve relevant Wikipedia articles corresponding to the 4k terminologies of Medical Encyclopedia.

\subsection{Global and local embeddings by \method} \label{sec:trial2vec_encode}
\modelname embeddings pretrained with MLM on clinical corpora still hold weak semantic meaning. Meanwhile, previous sentence embedding BERTs all take an average pooling over token embeddings, which causes the semantic meaning vanishing when applied to lengthy clinical trials. Therefore, we propose \method architecture that exploits the \textit{global} and \textit{local} embeddings for trial based on its meta-structure. 

We split the attributes of a trial into two distinct sets: \textit{key attributes} and \textit{contexts}. The first component includes the trial title, intervention, condition, and main measurement, which are sufficient to retrieve a pool of coarsely relevant trial candidates; the second includes descriptions, eligibility criteria, references, etc., which differentiate trials targeting similar diseases or interventions because they provide the multi-facet details regarding disease phases, study designs, targeted populations, etc. According to this design, local embeddings $\{\bv_{att}\}_{l=1}^L \in \mathbb{R}^{L\times D}$ are produced separately on each key attribute. On the other hand, a context embedding is obtained by encoding the context texts $\bv_{ctx} \in \mathbb{R}^D$. Note that the above encoding is all conducted by the same encoder. 

We further refine the local embeddings by context embeddings and aggregate them to yield the global trial embedding $\bv_{g}\in \mathbb{R}^D$. The refinement is performed by multi-head attention, as 
\begin{equation}
    \bv_g = \texttt{MultiHeadAttn}(\bv_{ctx},\{\bv_l\}_l^L, \bW),
\end{equation}
which relocates the attention over key attributes to enhance discrminative power of the yielded global embedding.

\subsection{Hierarchical contrastive learning}\label{sec:contrastive_learning}
For data-efficient contrastive learning, we utilize the meta-structure \& medical knowledge for contrasting local and global embeddings hierarchically. 

\noindent\textbf{Global contrastive loss.} The first objective is to maximize the semantic in trial embeddings for similarity search. Instead of doing in-batch instance-wise contrastive loss like SimCSE, we propose to sample informative negative pairs by exploiting the trial meta-structure. As shown by Fig. \ref{fig:overview}, some trials may be linked by a common attribute like disease or intervention. Denote a trial consisting of several attributes by
\begin{equation}
\bx = \{x^{\text{title}},x^{\text{intv}},x^{\text{dise}},x^{\text{out}},x^{\text{ctx}}\},
\end{equation}
we can build an informative negative sample by replacing its title with a trial which also targets for disease $x^{\text{dise}}$ by
\begin{equation}
    \bx^{-} = \{x^{\text{title}-},x^{\text{intv}},x^{\text{dise}},x^{\text{out}},x^{\text{ctx}}\}.
\end{equation}
Meanwhile, we apply a random attribute dropout towards $\bx$ to formulate a positive sample as
\begin{equation}
    \bx^{+} = \{x^{\text{title}},x^{\text{dise}},x^{\text{out}},x^{\text{ctx}}\}.
\end{equation}
InfoNCE loss is utilized in a batch of $B$ trials as
\begin{equation}
    \mathcal{L}_g =  - \sum_{i=1}^B \log \frac{\exp(\psi(\bv_{gi},\bv_{gi}^+))}{\sum_{v_{gi}^- \in \mathcal{V}_i^{-}} \exp(\psi(\bv_{gi},\bv_{gi}^-))},
\end{equation}
where the negative sample set $\mathcal{V}_{i}^- = \{\bv_{gi}^-\} \cup \{\bv_{gj}\}_{j\neq i}$; $\psi(\cdot,\cdot)$ measures the cosine similarity between two vectors. The global contrastive loss here encourages the model to capture the attribute of interest by discriminating the subtle differences of input trial attributes, which prevent the semantic meanings from vanishing due to the average pooling over all trial texts.

\noindent\textbf{Local contrastive loss.} In addition to the global trial embeddings, we put supervision on local embeddings to inject medical knowledge into the model. Unlike general texts, two medical texts can be overlapped word-wise dramatically but still describe two distinct things\footnote{For instance, replacing \textit{Olaparib} in "\textit{A Phase I, Open-Label, 2 Part Multicentre Study to Assess the Safety and Efficacy of Olaparib}" with another intervention like \textit{Vitamin D} renders a total different study topic.}, which is challenging for similarity computing. To strengthen \modelname discriminative power for medical texts, we extract key medical entities in each text as \footnote{Done by SciSpacy \url{https://scispacy.apps.allenai.org/}.}
\begin{equation}
    E(x^{att}) = \{e_1,e_2,e_3,e_4\},
\end{equation}
then a positive sample is built by mapping one entity $e_1$ to its canonical name or a similar entity under the same parental conception $\hat{e}_1$ defined by UMLS as
\begin{equation}
    E(x^{att+}) = \{\hat{e}_1,e_2,e_3,e_4\}.
\end{equation}
Similarly, negative sample is built by deletion or replacing one entity with another dissimilar one. InfoNCE loss is therefore used by
\begin{equation}
    \mathcal{L}_l = -\sum_{i=1}^B \log \frac{\psi(\bv_{li},\bv_{li}^+)}{\sum_{\mathcal{V}_{li}^-} \exp(\psi(\bv_{li},\bv_{li}^-)) }.
\end{equation}
We at last jointly optimize the global and contrastive losses as
\begin{equation}
    \mathcal{L} = \mathcal{L}_g + \mathcal{L}_l.
\end{equation}

\subsection{Application of global \& local embeddings} \label{sec:application}
The hierarchical contrastive learning offers extraordinary flexibility of \method for various downstream tasks in \textit{zero-shot} learning. At first, the global trial embeddings $\bv_g$ can be directly used for similarity search by comparing trial pair-wise cosine similarities. The computed trial embeddings can also help identify and discover research topics when we apply visualization techniques. On the other hand, we can also execute query search using partial attributes crediting to the contrastive learning between local and global embeddings. When we need do trial-level predictive tasks, e.g., trial termination prediction, a classifier can be attached to the pretrained global trial embeddings and learned; the backbone \modelname is also capable of offering short medical sentence retrieval because of local contrastive learning.

\section{Experiments}
In this section, we conduct five types of experiments to answer the following research questions:
\begin{itemize}[leftmargin=*, itemsep=0pt, labelsep=5pt]
    \item \textbf{Exp 1 \& 2.} How does \method perform in complete and partial retrieval scenarios?
    \item \textbf{Exp 3.}  How do the proposed SSL tasks / embedding dimension contribute to the performance?
    \item \textbf{Exp 4.} Is the trial embedding space interpretable and aligned with medical ontology?
    \item \textbf{Exp 5.} How useful do well-trained \method contribute to downstream tasks, e.g., trial outcome prediction, after fine-tuned?
    \item \textbf{Exp 6.} Qualitative analysis of the retrieval results and what are the differences of \method and baselines?
\end{itemize}

\begin{table}[t]
  \centering
  \caption{Statistics of trial status in \datawebsite database where we conclude \emph{Approved} \& \emph{Completed} as completion; \emph{Suspended}, \emph{Terminated}, and \emph{Withdrawn} as the termination for trial outcome prediction.}
  \resizebox{0.5\textwidth}{!}{%
    \begin{tabular}{lllll}
    \toprule
    Approved & Completed & Suspended & Terminated & Withdrawn \\
    174   & 210,237 & 1,658  & 22,208 & 10,439 \\
    Available & Enrolling & Unavailable & Not recruiting & Recruiting \\
    237   & 3,662  & 45,128 & 18,171 & 60,362 \\
    \midrule
    Completion & Termination & Summary & Others &  \\
    210,411 & 34,305 & 244,716 & 127,560 &  \\
    \bottomrule
    \end{tabular}%
    }
  \label{tab:status_stats}%
\end{table}%

\begin{table*}[t]
  \centering
  \caption{Precision/Recall and nDCG of the retrieval models on the labeled test set. Values in parenthesis show 95\% confidence interval.  Best values are in bold.}
  \resizebox{\textwidth}{!}{%
    \begin{tabular}{l|lllllll}
    \toprule
    Method & Prec@1 & Prec@2 & Prec@5 & Rec@1 & Rec@2 & Rec@5 & nDCG@5 \\
    \hline
    TF-IDF & 0.5132(0.063) & 0.4386(0.045) & 0.3828(0.057) & 0.1871(0.038) & 0.3172(0.026) & 0.6147(0.044) & 0.5480(0.034)\\
    BM25 & 0.7015(0.044) & 0.5640(0.041) & 0.4246(0.032) & 0.3358(0.038) & 0.4841(0.050) & 0.7666(0.031) & 0.7312(0.033) \\
    Word2Vec & 0.7492(0.071) & 0.6476(0.044) & 0.4712(0.033) & 0.3008(0.054) & 0.4929(0.042) & 0.7939(0.041) & 0.7712(0.032) \\
    BERT & 0.7264(0.050) & 0.6219(0.060) & 0.4324(0.027) & 0.3257(0.051) & 0.4896(0.054) & 0.7611(0.041) & 0.7370(0.047) \\
    BERT$_{\text{Whiten}}$ & 0.7476(0.094) & 0.6630(0.045) & 0.4525(0.029) & 0.3672(0.045) & 0.5832(0.042) & 0.8355(0.021) & 0.8129(0.024) \\
    BERT$_{\text{SimCSE}}$ & 0.6788(0.039) & 0.5995(0.035) & 0.4714(0.021) & 0.2824(0.034) & 0.4566(0.035) & 0.8098(0.025) & 0.7308(0.038) \\
    MonoT5$_{\text{Med}}$ & 0.6799(0.068) & 0.5810(0.061) & 0.4439(0.051) & 0.2904(0.032) & 0.4657(0.049) & 0.7570(0.037) & 0.7171(0.043) \\
    \method & \textbf{0.8810(0.026)} & \textbf{0.7912(0.049)} & \textbf{0.5055(0.039)} & \textbf{0.4216(0.046)} & \textbf{0.6465(0.060)} & \textbf{0.8919(0.030)} & \textbf{0.8825(0.029)}\\
    \bottomrule
    \end{tabular}%
    }
  \label{tab:complete_retrieval}%
\end{table*}%

\subsection{Dataset \& Setup}
\textbf{Trial Similarity Search.} We created a labeled trial dataset to evaluate the retrieval performance where paired trials are labeled as relevant or not. We keep 311,485 interventional trials from the total 399,046 trials. We uniformly sample 160 trials as the query trials. To overcome the sparsity of relevance, we take advantage of TF-IDF \cite{salton1983extended} to retrieve ranked top-10 trials as the candidate to be labeled, resulting in 1,600 labeled pairs of clinical trials. Unlike general documents, the clinical trial document contains many medical terms and formulations. We recruited clinical informatics researchers, and each is assigned 400 pairs to label as relevant or not using label $\{1,0\}$. To keep labeling processes in line, we specify the minimum annotation guide for judging relevance: (1) same disease; or (2) same intervention and similar diseases (e.g., cancer on distinct body parts).
We use precision@k (prec@k), recall@k (rec@k), and nDCG@5 to evaluate and report performances.
\begin{align}
    prec@k & = \frac{\text{\# of relevant trials in the top k results}}{k}, \label{eq:precision_k}\\
    rec@k & = \frac{\text{\# of relevant trials in the top k results}}{\text{\# of relevant trials in all candidate trials}}.
\end{align}

\noindent \textbf{Trial termination prediction.} We can take the pretrained \method embeddings for predicting the trial outcomes, i.e., if the trial will be terminated or not. We add one additional fully-connected layer on the tail of \method. The targeted outcomes are in the status section of clinical trials, described by Table \ref{tab:status_stats}. We formulate the outcome prediction as a binary classification problem to predict the \textit{Completion} or \textit{Termination} of trials where we get 210,411 and 34,305 trials as positive and negative labeled, respectively. We take 70\% of all as the training set and 20\% as the test set; the remaining 10\% is used as the validation set for tuning and early stopping. We utilize three metrics for evaluation: accuracy (ACC), area under the Receiver Operating Characteristic (ROC-AUC), and area under Precision-Recall curve (PR-AUC).

\subsection{Baselines \& Implementations}
We take the following baselines for retrieval: TF-IDF   \cite{salton1983extended,salton1988term}, BM25 \cite{trotman2014improvements}, Word2Vec \cite{mikolov2013efficient}, BERT-Whitening \cite{huang2021whiteningbert,su2021whitening},  BERT-SimCSE \cite{gao2021simcse}, and MonoT5 \cite{kirk2021overview,pradeep2022neural}. Details of these methods can be seen in Appendix \ref{appx:2}.

We keep all methods' embedding dimensions at 768. We start from a BERT-base model to continue pre-training on clinical domain corpora, yielding our \modelname, which supports as the backbone for BERT-Whitening and BERT-SimCSE for fair comparison. We take 5 epochs with batch size 100 and the learning rate 5e-5. In the second SSL training phase, AdamW optimizer with a learning rate of 2e-5, batch size of 50, and weight decay of 1e-4 is used. Experiments were done with 6 RTX 2080 Ti GPUs. 

\begin{figure}[t]
    \centering
    \includegraphics[width=0.95\linewidth]{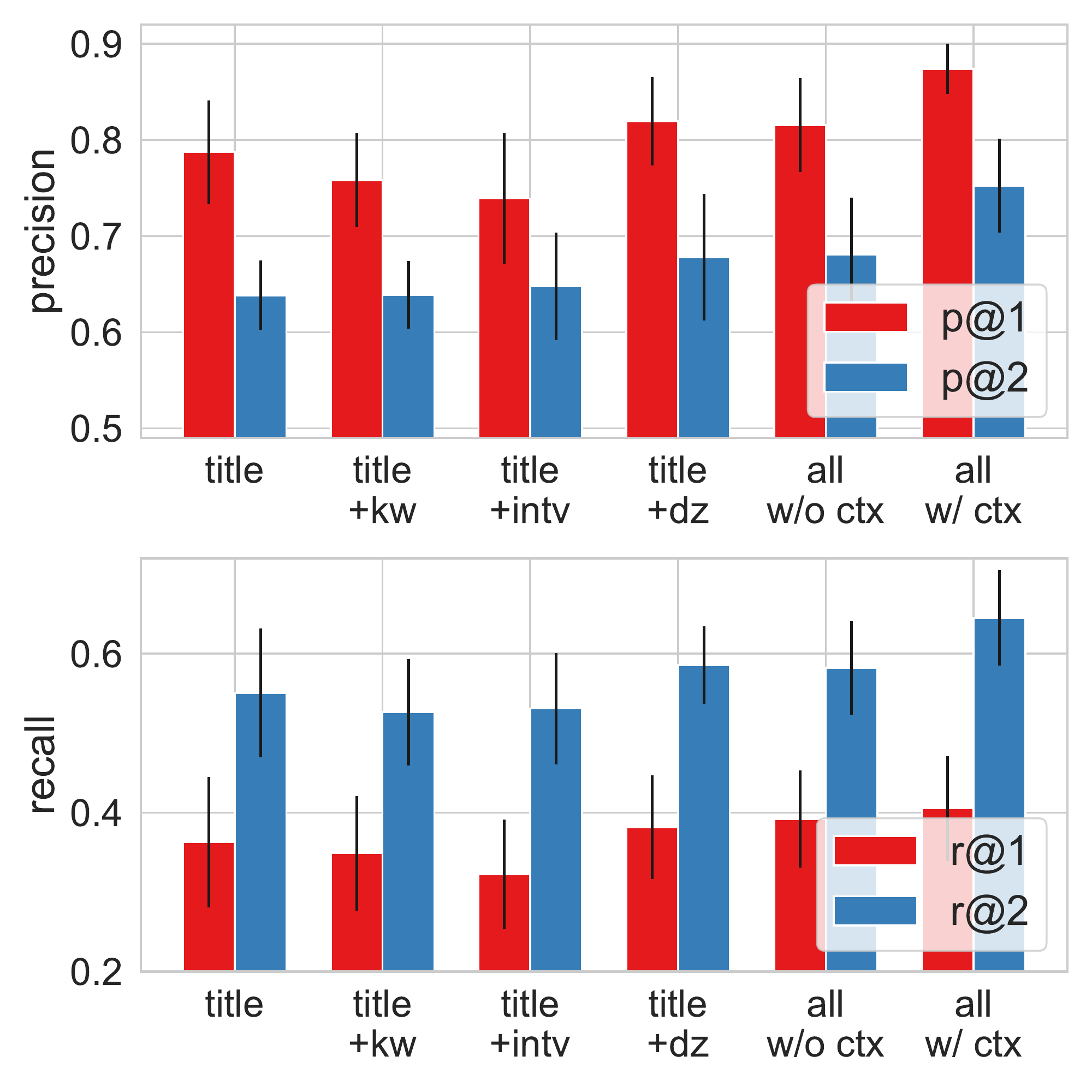}
    \caption{Performance of \method on the partial retrieval scenarios. 
    We use a different part of the trial as queries to retrieve similar trials, including keyword
    \textit{kw}, intervention \textit{intv}, disease \textit{dz}, context \textit{ctx}.  Error bars indicate the 95\% confidence interval of results.}
    \label{fig:partial_retrieval}
\end{figure}

\begin{figure*}[t]
\centering
\begin{minipage}[t]{0.48\textwidth}
\centering
\includegraphics[width=0.8\textwidth]{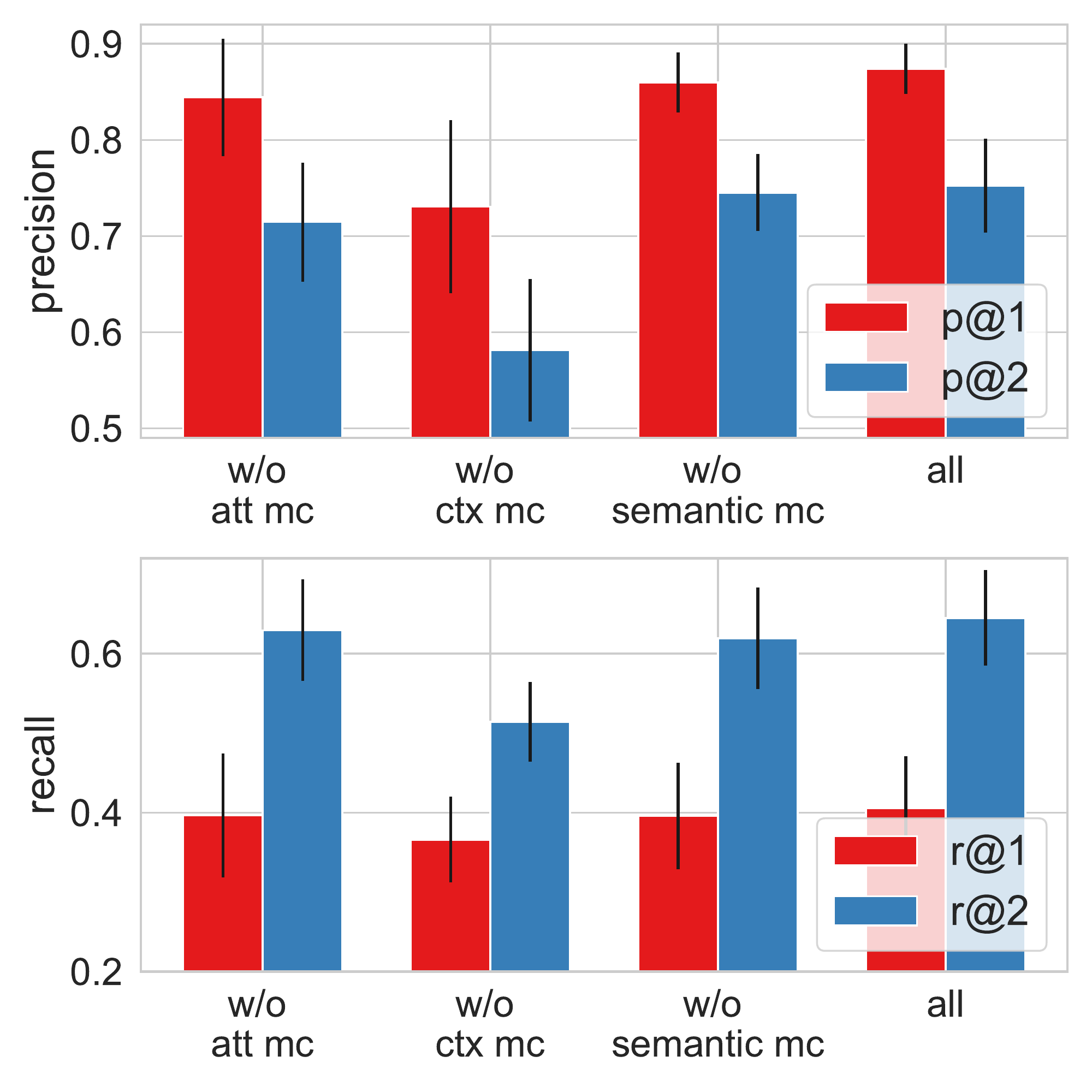}
\caption{Ablation study on the contribution of each Task to the final result. \textit{att}, \textit{mc}, \textit{ctx} are short for attribute, matching, context, respectively. \textit{all} indicate the full \method that all tasks are used. \label{fig:ablation_ssl}}
\end{minipage}
\hfill
\begin{minipage}[t]{0.48\textwidth}
\centering
\includegraphics[width=0.8\textwidth]{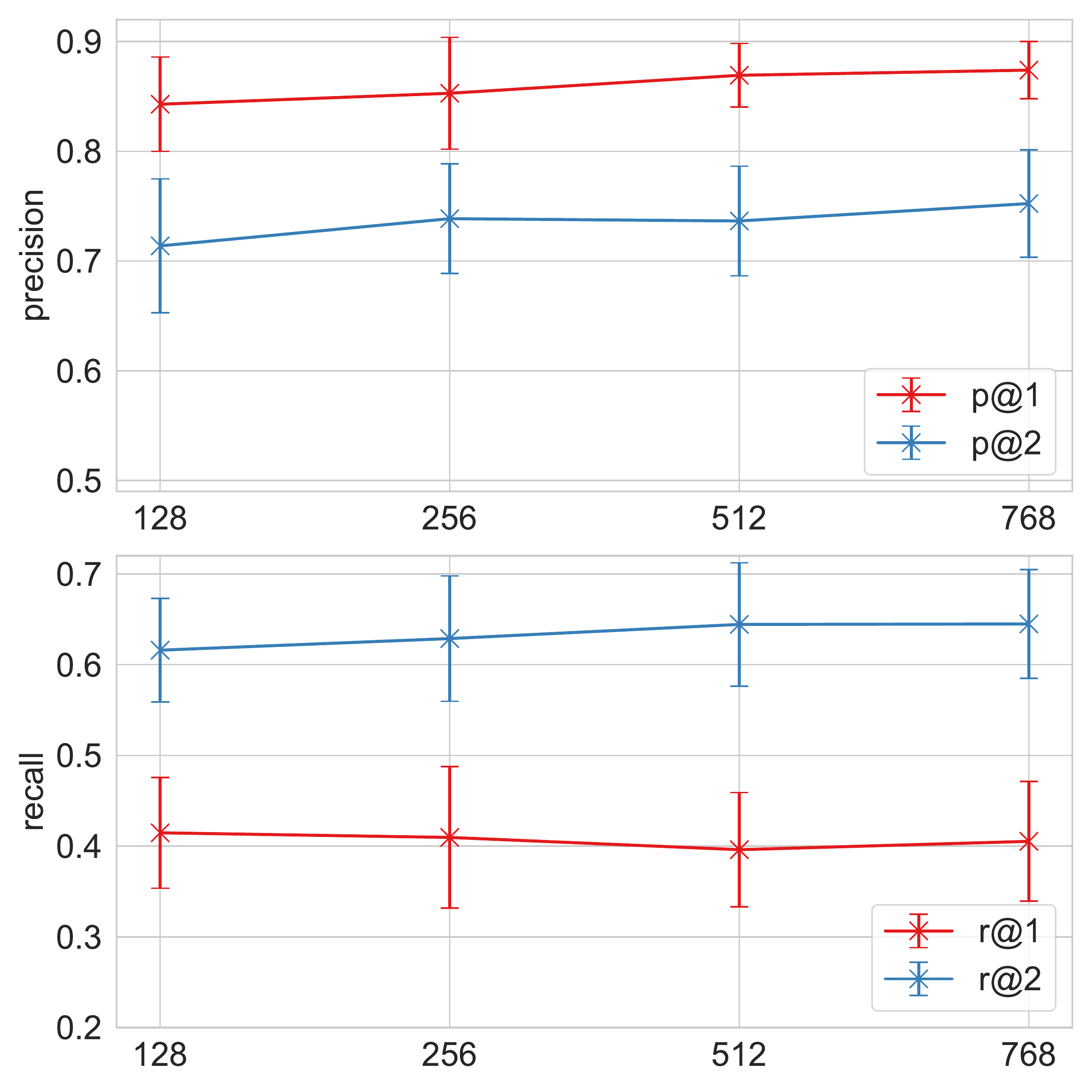}
\caption{Analysis of the influence of embedding dimensions on retrieval quality by \method: embedding dim in 128, 256, 512, 768. Error bars show the 95\% confidence interval. \label{fig:emb_dim}}
\end{minipage}
\end{figure*}

\begin{figure*}[t]
    \centering
    \includegraphics[width=0.99\linewidth]{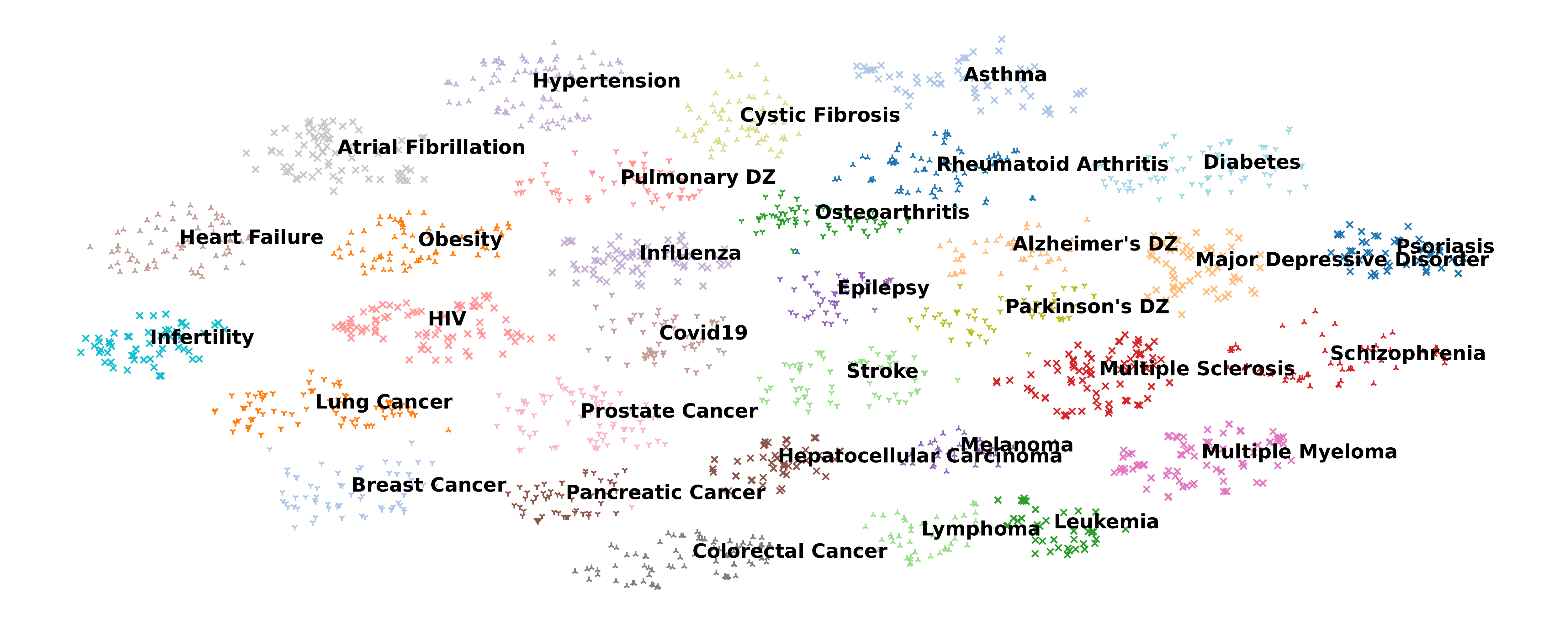}
    \caption{2D visualization of the trial-level embeddings obtained by \method (dimension reduced by t-SNE). It can be seen trials are automatically classified into clusters by topic (diseases) in the embedding space. For example, a series of tumor-related trials (e.g., Breast and Pancreatic Cancers) are on the bottom of the embedding space.}
    \label{fig:emb_vis_by_disease}
    \vspace{-1em}
\end{figure*}

\subsection{Exp 1. Complete Trial Similarity Search}
Since labels are unavailable in the training phase, we only chose unsupervised/self-supervised baselines. Results are shown by Table ~\ref{tab:complete_retrieval}. \method outperforms all baselines with a great margin. It has around 15\% improvement on each metrics than the best baselines on average. For baselines, all except for TF-IDF have similar performance. When $k$ is small, the precision gap between \method and baselines is large; when $k$ is large, all methods encounter precision reduction. That is because the pool of candidate trials are 10 but the number of positive pairs for each are often less than 5, which limits the maximum of the numerator of $prec@k$ in Eq. \eqref{eq:precision_k}. Likewise, \method also shows stronger performance in $rec@k$ because it is discounted by the maximum number of positive pairs.

\begin{table}[t]
  \centering
  \caption{Trial outcome prediction performances of baselines and \method, after fine-tuned.}
  \resizebox{0.5\textwidth}{!}{%
    \begin{tabular}{l|lll}
    \toprule
    Method & ACC   & ROC-AUC & PR-AUC \\
    \hline
    TF-IDF & 0.8571(0.002) & 0.7194(0.004) & 0.2960(0.008) \\
    Word2Vec & 0.8574(0.002) & 0.7189(0.005) & 0.2906(0.007) \\
    TrialBERT  & 0.8559(0.002) & 0.7277(0.006) & 0.3109(0.006) \\
    \method & \textbf{0.8622(0.002)} & \textbf{0.7332(0.004)} & \textbf{0.3137(0.007)} \\
    \bottomrule
    \end{tabular}%
    }
  \label{tab:outcome_pred_res}%
\end{table}%

\begin{table*}[t]
  \centering
  \caption{Case studies comparing the retrieval performance of the \method with baseline models. Due to the space limits, only title and NCT ID of trials are given.}
  \resizebox{\textwidth}{!}{%
    \begin{tabular} {p{13em}|p{13em}p{13em}p{13em}}
    \toprule
    Query Trial & TF-IDF & TrialBERT & \method 
    \bigstrut\\
    \hline
    [NCT02972294] HiFIT Study : Hip Fracture: Iron and Tranexamic Acid (HiFIT)   & [NCT01221389] Study Using Plasma for Patients Requiring Emergency Surgery (SUPPRES)
 & [NCT04744181] Patient Blood Management In CARdiac sUrgical patientS (ICARUS) & [NCT01535781] Study of the Effect of Tranexamic Acid Administered to Patients With Hip Fractures. Can Blood Loss be Reduced? \bigstrut[t]\\
 
    [NCT01590342] Diclofenac for Submassive PE (AINEP-1) & [NCT04006145] A Phase 2 Study of Elobixibat in Adults With NAFLD or NASH & [NCT04156854] Intravascular Volume Expansion to Neuroendocrine-Renal Function Profiles in Chronic Heart Failure
 & [NCT00247052] Non Steroidal Anti Inflammatory Treatment for Post Operative Pericardial Effusion \\
    \bottomrule
    \end{tabular}%
    }
  \label{tab:casestudies}%
\end{table*}%

Interestingly, the state-of-the-art sentence BERTs, e.g., BERT-whitening and BERT-simCSE, have limited improvement over original BERT and even Word2Vec. Unlike general documents, clinical trials may be overlapped in much content but still be irrelevant if the key entities are different. This special characteristic causes the assumption of \textit{a document with similar passage is relevant} \cite{craswell2020overview} used in general document retrieval but invalidated in clinical trial retrieval. Without well-designed SSL, it is hard for these methods to learn these subtle differences. Moreover, clinical trial documents are often much longer than the general documents in those open datasets. There are 622.4 words per trial on average, while the general STS benchmark has below 15 words per sample, e.g., STS-12: 10.8, STS-13: 8.8, STS-14: 9.1, etc \cite{cer2017semeval}.  We also observed the simple negative sampling strategy of SimCSE is insufficient to learn effective long document embeddings. In comparison, \method leverages the meta-structure of clinical trials to focus on the most informative attributes, with additional context-based refinement, producing embeddings superior in semantic representation.   

\subsection{Exp 2. Partial Query Trial Retrieval}
We further investigate the partial trial retrieval scenario where users intend to find similar trials with short and incomplete descriptions, e.g., partial attributes. Results are illustrated by Fig~ \ref{fig:partial_retrieval}. We start by measuring how well \method only utilizes the title for trial retrieval. It is witnessed that using title is sufficient to yield comparable performance as the best baseline for complete retrieval shown in Table \ref{tab:complete_retrieval}.  Nonetheless, we identify that concatenating keywords or intervention with the title reduces performance. Combining title and disease yields similar performance as involving all attributes. This phenomenon signifies that the disease plays a vital role in trial similarity and is always recommended to be involved in query trial retrieval.

\subsection{Exp 3. Ablation Studies}
We conducted ablation studies to measure how SSL tasks and embedding dimensions contribute to final results. Results are shown by Fig. \ref{fig:ablation_ssl}, where we remove one Task for each setting and reevaluate. Here, \textit{att mc} and \textit{ctx mc} corresponds to the global contrastive loss by negative sampling on key attributes and contexts, respectively; \textit{semantic mc} indicates the local contrastive loss. We observe that ctx mc is very important. Without it, only attributes of trials are included in the training and inference of \method, thus resulting in a significant performance drop. However, even only using a small segment of trials (the attributes), \method still reaches similar performance as BERT-SimCSE that receives the whole trial document as inputs. This demonstrates the importance of picking high-quality negative samples during the CL process. Similarly, we observe other two tasks also improve the retrieval quality.

Fig. \ref{fig:emb_dim} illustrates the retrieval performance on different embedding dimensions. We identify that reducing embedding dimension does not affect the performance of \method much, i.e., one can choose a small embedding dimension (e.g., 128) without suffering much performance degradation while saving lots of storage and computational resources.

\subsection{Exp 4. Embedding Space Visualization}
Fig. \ref{fig:emb_vis_by_disease} plots the 2D visualization of the embedding space of \method using t-SNE \cite{van2008visualizing} where around 2k trials uniformly sampled from 300k trials. The tag texts illustrate the target diseases of trials with different colors. We observe that these trials embeddings show interpretable clusters corresponding to target disease categories. More discussions about this visualization can be referred to Appendix \ref{appx:1}.

\subsection{Exp 5. Trial Termination Prediction}
Results are illustrated by Table \ref{tab:outcome_pred_res}. Compared with the shallow models, BERT-based methods gain better performance, which credits the deep architecture of transformers with stronger learning capability. \method takes a hierarchical encoding for trial documents on meta-structure thus better revealing the trial characteristics, which plays a central role in predicting its potention outcomes.

\subsection{Exp 6. Case Study}
We perform a qualitataive analysis of similarity search results and two baselines. Results are shown in Table \ref{tab:casestudies}. These two case studies show that TF-IDF and BERT models all tend to put attention on frequent words in query trials, e.g., \textit{blood} and \textit{iron} in case study 1; and \textit{heart failure} in case study 2. This bias comes from the average pooing taken onto all token embeddings.  The top-1 relevant clinical trial retrieved by \method, on the other hand, provides a more similar trial thanks to the hierarchical encoding and specific local and global contrastive learning.  We add more explanations regarding these cases in Appendix \ref{appx:3}.

\section{Conclusion}
This paper investigated utilizing BERT with self-supervision for encoding trial into dense embeddings for similarity search. Experiments show our method can succeed in zero-shot trial search under various settings. The embeddings are also useful for trial downstream predictive tasks. The qualitative analysis, including embedding space visualization and case studies, further verifies that \method gets a medically meaningful understanding of clinical trials.

\section*{Limitations}
The empirical evaluation of this method is mainly done on the clinical trial documents drawn from \datawebsite which were fully written in English. It might be the best fit when this method is applied to documents in other languages. Although we have tried our best to collect trial relevance datasets, it is still possible that the datasets used for evaluation are not able to cover all cases.

The proposed framework encodes trial documents into compact embeddings for search. It encounters failure cases some time as wrong trials are retrieved. It should be used with discretion when applied to clinical trial research or by individual volunteers who intend to look for trials research. Retrieved results in practice should be used under the supervision with professional clinicians.


\bibliography{anthology,custom}
\bibliographystyle{acl_natbib}

\clearpage

\appendix
\begin{table}[t]
  \centering
  \caption{List of text corpora used for continual pretraining of \modelname.}
   \resizebox{0.4\textwidth}{!}{
    \begin{tabular}{ll}
    \hline
    Corpus & Number of words \\
    \midrule
    ClinicalTrials.gov & 240M \\
    Medical Encyclopedia & 3M \\
    Wikipedia Articles & 11M \\
    \bottomrule
    \end{tabular}%
    }
  \label{tab:pretraining_corpus}%
\end{table}%

\section{Baselines for clinical trial similarity search} \label{appx:2}
\begin{itemize}
    \item TF-IDF \cite{salton1983extended,salton1988term}. It is short for term frequency–inverse document frequency that has been widely used for information retrieval systems for decades. One can use TF-IDF for document retrieval by concatenating scores of all words in this document then computing cosine distance between document vectors.
    \item BM25 \cite{trotman2014improvements}. A bag-of-words retrieval method commonly used in practice. We run it based on the rank-bm25 package \footnote{\url{https://pypi.org/project/rank-bm25/}} with its default hyperparameters.
    \item Word2Vec \cite{mikolov2013efficient}. It is a classic dense retrieval method by building distributed word representations by self-supervised learning methods (CBOW). We take an average pooling of word representations in a document for retrieval by cosine distance. We use gensim \footnote{\url{https://radimrehurek.com/gensim/models/doc2vec.html}} to run this method.
    \item BERT. We take an average pooling over all token embeddings at the last layer of it for similarity computation. We take the TrialBERT pretrained on all the clinical trial documents.
    \item BERT-Whitening \cite{huang2021whiteningbert,su2021whitening}. This is an unsupervised post-processing method that uses anisotropic BERT embeddings \cite{ethayarajh2019contextual,li2020sentence} to improve semantic search. We take the average of last and first layer of its BERT embeddings following \citet{su2021whitening}.
    \item BERT-SimCSE \cite{gao2021simcse}. It is a contrastive sentence representation learning method stemming from InfoNCE loss. It simply takes other samples in batch as negative samples.
    \item MonoT5-Med \cite{pradeep2022neural}. It was proposed in \cite{kirk2021overview} for matching patient descriptive texts and clinical trial documents via T5 model \cite{raffel2020exploring} based on prompts. We use its version finetuned on Med Marco dataset \cite{koopman2016test}.
\end{itemize}

\section{Embedding space visualization} \label{appx:1}
From Fig. \ref{fig:emb_vis_by_disease}, trial embeddings are clearly clustered into topics with self-supervised learning, which provides a great help for topic mining and discovery for the existing clinical trials. For instance, we can find that cancers that happen on different body parts are near to each other on the bottom of the embedding space (Prostate Cancer, Breast Cancer, Pancreatic Cancer, Colorectal Cancer, etc.). Also, the diseases which are related to brain function, e.g., Alzheimer's Disease, Parkinson's Disease, Major Depressive Disorder, etc.  Other examples include Covid19, Influenza, Pulmonary Disease, etc. 

The reason is that we explicitly utilize the knowledge from attributes of trials for negative sample building, which endows the embedding space the ability to discriminate trials' similarity. These similar trials can also have similar characteristics like having similar recruiting criteria or targeting similar outcome measures, which are captured by \method by refining the embeddings of attributes by detailed descriptions.  Based on this observation, we can infer that such medically meaningful trial embeddings would be beneficial to downstream tasks on clinical trials, e.g., trial outcome prediction.

\section{Case Study} \label{appx:3}
For the first case, the query trial is [NCT02972294], which studies using Tranexamic acid and Iron Isomaltoside to reduce the occurrence of Anemia and blood transfusion in hip fracture cases.  We show the top-1 retrieved by three methods on the right. Trial found by TF-IDF studies the efficiency of plasma in patients with Hemorrhagic shock; BioBERT finds a trial about patients undergoing heart surgery who have Anaemia to test if a correction of iron reduces red blood cell transfusion requirements. \method finds a trial that studies Tranexamic acid effect in blood loss in hip fracture operations. \method result is highly relevant to the query trial as it has the identical drug on blood loss of the same type of operation.

In the second example, the query trial tries to investigate the benefits of Diclofenac for Normotensive patients with acute symptomatic Pulmonary Embolism and Right Ventricular Dysfunction. TF-IDF finds an irrelevant study on the efficacy and safety of Elobixibat for adults with NAFLD or NASH. 
\modelname also retrieves an irrelevant study on Intravascular Volume Expansion to Neuroendocrine-Renal Function Profiles in Chronic Heart Failure. On the other hand, \method digs out a trial that studies the same type of drug with a similar purpose as the target's: evaluating the efficiency of NSAID (Diclofenac) to the evolution of postoperative (cardiac surgery) pericardial effusion.

\end{document}